\pdfoutput=1
\documentclass[runningheads]{llncs}
\usepackage{graphicx}
\usepackage{comment}
\usepackage{amsmath,amssymb} 
\usepackage{color}
\usepackage{multirow}
\usepackage[colorlinks]{hyperref}

\begin{document}
\pagestyle{headings}
\mainmatter
\def\ECCVSubNumber{1342} 

\title{Funnel Activation for Visual Recognition} 

\titlerunning{Funnel Activation for Visual Recognition}
%
\author{Ningning Ma\inst{1} \and
Xiangyu Zhang\inst{2}\thanks{Corresponding author}\and
Jian Sun\inst{2} }
\authorrunning{Ningning Ma et al.}
%
\institute{Hong Kong University of Science and Technology\and MEGVII Technology\\
\email{nmaac@cse.ust.hk, \{zhangxiangyu,sunjian\}@megvii.com}}
\maketitle

\newcommand{\bX}{\mathbf{X}}
\newcommand{\bT}{\mathbb{T}}
\newcommand{\red}[1]{{\color{red} #1}}

\begin{abstract}

We present a conceptually simple but effective funnel activation for image recognition tasks,
called \textit{Funnel activation (FReLU)}, that extends ReLU and PReLU to a 2D activation by adding a negligible overhead of spatial condition.
The forms of ReLU and PReLU are $y=max(x,0)$ and $y=max(x,px)$, respectively, while FReLU is in the form of $y=max(x, \bT(x))$, where $\bT(\cdot)$ is the 2D spatial condition. 
Moreover, the spatial condition achieves a pixel-wise modeling capacity in a simple way, capturing complicated visual layouts with regular convolutions.
We conduct experiments on ImageNet, COCO detection, and semantic segmentation tasks, showing great improvements and robustness of FReLU in the visual recognition tasks. Code is available at \url{https://github.com/megvii-model/FunnelAct}. 
\end{abstract}

\section{Introduction}

Convolutional neural networks (CNNs) have achieved state-of-the-art performance in many visual recognition tasks, such as image classification, object detection, and semantic segmentation.
As popularized in the CNN framework, one major kind of layer is the convolution layer, another is the non-linear activation layer.

First in the convolution layers, capturing the spatial dependency adaptively is challenging, many advances in more complex and effective convolutions have been proposed to grasp the local context adaptively in images \cite{dai2017deformable,holschneider1990real}.
The advances achieve great success especially on dense prediction tasks (e.g., semantic segmentation, object detection).
Driven by the advances in more complex convolutions and their less efficient implementations, a question arises: \textit{Could regular convolutions achieve similar accuracy, to grasp the challenging complex images?}

Second, usually right after capturing spatial dependency in a convolution layer $linearly$, then an activation layer acts as a scalar non-linear transformation.
Many insightful activations have been proposed \cite{maas2013rectifier,he2015delving,clevert2015fast,klambauer2017self}, but improving the performance on visual tasks is challenging, therefore currently the most widely used activation is still the Rectified Linear Unit (ReLU) \cite{nair2010rectified}.
Driven by the distinct roles of the convolution layers and activation layers, another question arises: 
\textit{Could we design an activation specifically for visual tasks?}

\begin{figure}[t]
\centering
\includegraphics[height=5cm]{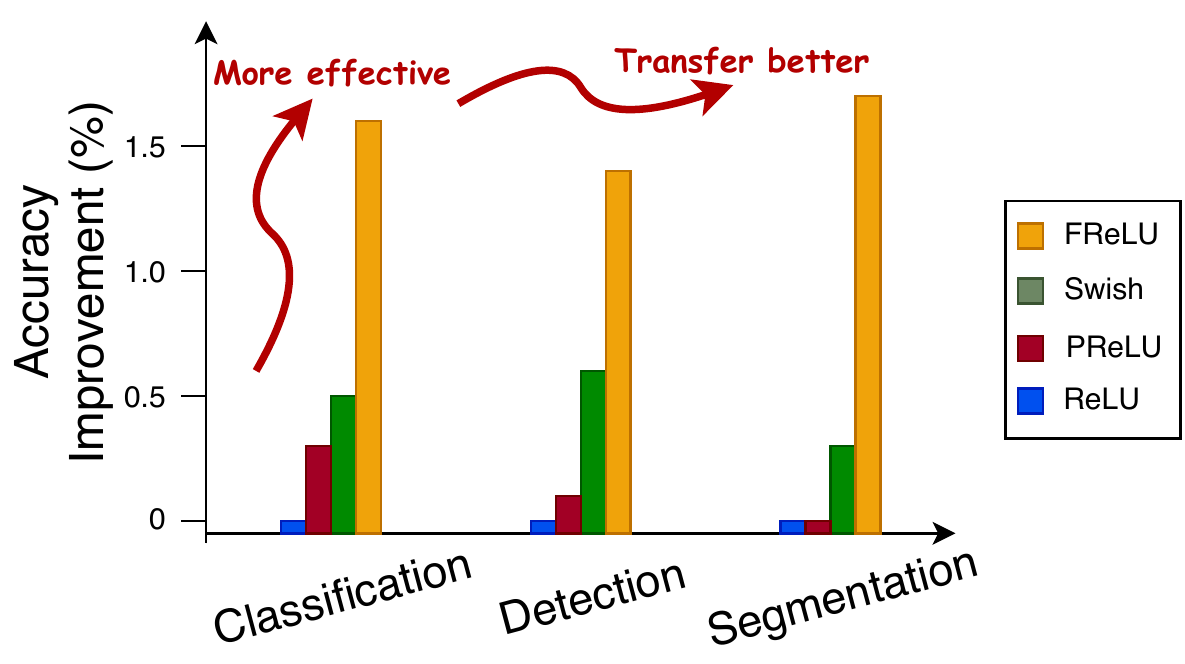}
\caption{\textbf{Effectiveness} and \textbf{generalization} performance.
We set the ReLU network as the baseline, and show the \textit{relative improvement} of accuracy on the three basic tasks in computer vision: image classification (Top-1 accuracy), object detection (mAP), and semantic segmentation (mean\_IU).
We use the ResNet-50 \cite{he2016deep} as the backbone pre-trained on the ImageNet dataset, to evaluate the generalization performance on COCO and CityScape datasets.
FReLU is more effective, and transfer better on all of the three tasks.}
\label{fig:result}
\end{figure}

To answer both questions raised above, we show that the simple but effective visual activation, together with the regular convolutions, can also achieve significant improvements on both dense and sparse predictions (e.g. image classification, see Fig. \ref{fig:result}).
To achieve the results, we identify spatially insensitiveness in activations as the main obstacle impeding visual tasks from achieving significant improvements and propose a new visual activation that eliminates this barrier.
In this work, we present a simple but effective visual activation that extends ReLU and PReLU to a 2D visual activation.

Spatially insensitiveness is addressed in modern activations for visual tasks.
As popularized in the ReLU activation, non-linearity is performed using a $max(\cdot)$ function, the condition is the hand-designed $zero$, thus in the scalar form: $y=max(x,0)$.
The ReLU activation consistently achieves top accuracy on many challenging tasks.
Through a sequence of advances \cite{maas2013rectifier,he2015delving,clevert2015fast,klambauer2017self}, many variants of ReLU modify the condition in various ways and relatively improve the accuracy.
However, further improvement is challenging for visual tasks.

Our method, called \textbf{Funnel activation (FReLU)}, extends the spirit of ReLU/PReLU by adding a spatial condition (see Fig. \ref{fig:framework}) which is simple to implement and only adds a negligible computational overhead.
Formally, the form of our proposed method is $y=max(x, \bT(x))$, where $\bT(x)$ represents the simple and efficient spatial contextual feature extractor.
By using the spatial condition in activations, it simply extends ReLU and PReLU to a visual parametric ReLU with a pixel-wise modeling capacity.





Our proposed visual activation acts as an efficient but much more effective alternative to previous activation approaches.
To demonstrate the effectiveness of the proposed visual activation, we replace the normal ReLU in classification networks, and we use the pre-trained backbone to show its generality on the other two basic vision tasks: object detection and semantic segmentation.
The results show that FReLU not only improves performance on a single task but also transfers well to other visual tasks.

\section{Related Work}


\subsubsection{Scalar activations}

Scalar activations are activations with single input and single output, in the form of $y=f(x)$.
The Rectified Linear Unit (ReLU) \cite{hahnloser2000digital,jarrett2009best,nair2010rectified} is the most widely used scalar activation on various tasks \cite{krizhevsky2012imagenet,simonyan2014very}, in the form of $y=max(x,0)$.
It is simple and effective for various tasks and datasets.
To modify the negative part, many variants have been proposed, such as Leaky ReLU \cite{maas2013rectifier}, PReLU \cite{he2015delving}, ELU \cite{clevert2015fast}.
They keep the positive part identity and make the negative part dependent on the sample adaptively.

Other scalar methods such as the sigmoid non-linearity has the form $\sigma(x)=1/(1+e^{-x})$, and the Tanh non-linearity has the form $tanh(x)=2\sigma(2x)-1$.
These activations are not widely used in deep CNNs mainly because they saturate and kill gradients, also involve expensive operations (exponentials, etc.).

Many advances followed \cite{klambauer2017self,singh2019filter,agostinelli2014learning,hendrycks2016bridging,qiu2018frelu,elfwing2018sigmoid,xu2015empirical}, and recent searching technique contributes to a new searched scalar activation called Swish \cite{ramachandran2017searching} by combing a comprehensive set of unary functions and binary functions.
The form is $y=x * Sigmoid(x)$, outperforms other scalar activations on some structures and datasets, and many searched results show great potential.

\subsubsection{Contextual conditional activations}
Besides the scalar activation which only depends on the neuron itself, conditional activation is a many-to-one function, which activates the neurons conditioned on contextual information.
A representative method is Maxout \cite{goodfellow2013maxout}, it extends the layer to a multi-branch and selects the maximum.
Most activations apply a non-linearity on the linear dot product between the weights and the data, which is: $f(w^Tx+b)$.
Maxout computes the $max(w^T_1x+b_1,w^T_2x+b_2)$, and generalizes ReLU and Leaky ReLU into the same framework.
With dropout \cite{hinton2012improving}, the Maxout network shows improvement.
However, it increases the complexity too much, the numbers of parameters and multiply-adds has doubled and redoubled.

Contextual gating methods \cite{dauphin2017language,wu2016multiplicative} use contextual information to enhance the efficacy, especially on RNN based methods, because the feature dimension is relatively smaller.
There are also on CNN based methods \cite{van2016conditional}, since 2D feature size has a large dimension, the method is used after a feature reduction.

The contextually conditioned activations are usually channel-wise methods.
However, in this paper, we find the spatial dependency is also important in the non-linear activation functions.
We use light-weight CNN technique depth-wise separable convolution to help with the reduction of additional complexity.

\subsubsection{Spatial dependency modeling}
Learning better spatial dependency is challenging, 
Some approaches use different shapes of convolution kernels \cite{szegedy2015going,szegedy2016rethinking,szegedy2017inception} to aggregate the different ranges of spatial dependences.
However, it requires a multi-branch that decreases efficiency.
Advances in convolution kernels such as atrous convolution \cite{holschneider1990real} and dilated convolution \cite{yu2015multi} also lead to better performance by increasing the receptive field.

Another type of methods learn the spatial dependency adaptively, such as STN \cite{jaderberg2015spatial}, active convolution \cite{jeon2017active}, deformable convolution \cite{dai2017deformable}.
These methods adaptively use the spatial transformations to refine the short-range dependencies, especially for dense vision tasks (e.g. object detection, semantic segmentation).
Our simple FReLU even outperforms them without complex convolutions.

Moreover, the non-local network provides the methods to capture long-range dependencies to address this problem.
GCNet \cite{cao2019gcnet} provides a spatial attention mechanism to better use the spatial global context.
Long-range modeling methods achieve better performance but still require additional blocks into the origin network structure, which decreases efficiency.  
Our method address this issue in the non-linear activations, solve this issue better and more efficiently.

\subsubsection{Receptive field}
The region and size of receptive field are essential in vision recognition tasks \cite{zhou2014object,noh2017large}.
The work on effective receptive field \cite{luo2016understanding,glorot2010understanding} finds that different pixels contribute unequally and the center pixels have a larger impact.
Therefore, many methods have been proposed to implement the adaptive receptive field \cite{dai2017deformable,zhu2019deformable,zhao2018psanet}.
The methods achieve the adaptive receptive field and improve the performance, by involving additional branches in the architecture, such as developing more complex convolutions or utilizing the attention mechanism.
Our method also achieves the same goal, but in a more simple and efficient manner by introducing the receptive field into the non-linear activations. 
By using the more adaptive receptive field, we can approximate the layouts in common complex shapes, thus achieve even better results than the complex convolutions, by using the efficient regular convolutions.

\section{Funnel Activation}

\begin{figure}[t]
\centering
\includegraphics[height=6cm]{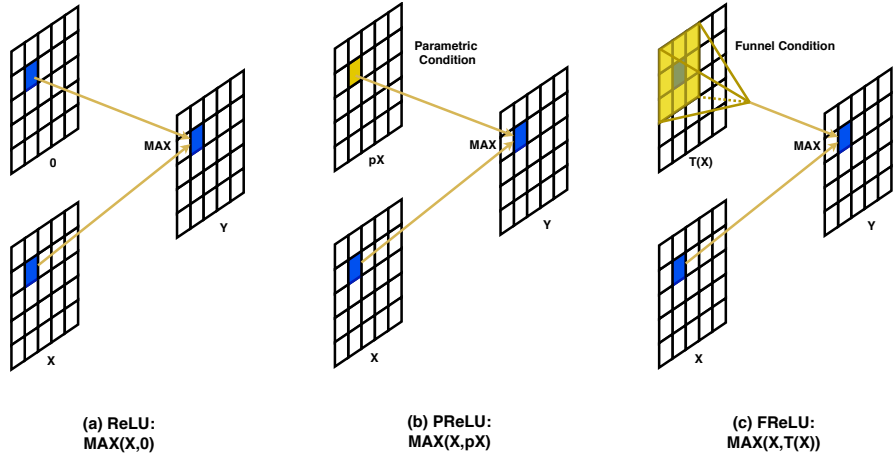}
\caption{\textbf{Funnel activation.} We propose a novel activation for visual recognition we call $FReLU$  that follows the spirit of ReLU/PReLU and extends them to 2D by adding a visual funnel condition $\bT(x)$. (a) ReLU with a condition zero; (b) PReLU with a parametric condition; (c) FReLU with a visual parametric condition.}
\label{fig:framework}
\end{figure}

FReLU is designed specifically for visual tasks and is conceptually simple: the condition is a hand-designed zero for ReLU and a parametric $px$ for PReLU, to this we modify it to a 2D funnel-like condition dependent on the spatial context. The visual condition helps extract the fine spatial layout of an object.
Next, we introduce the key elements of FReLU, including the funnel condition and the pixel-wise modeling capacity, which are the main missing parts in ReLU and its variants. 

\subsubsection{ReLU}
We begin by briefly reviewing the ReLU activation.
ReLU, in the form $max(x,0)$, uses the $max(\cdot)$ to serve as non-linearity and uses a hand-designed $zero$ as the condition.
The non-linear transformation acts as a supplement of the linear transformation such as convolution and fully-connected layers.

\subsubsection{PReLU}
As an advanced variant of ReLU, PReLU has an original form $max(x,0)+p\cdot min(x,0)$, where $p$ is a learnable parameter and initialized as 0.25. However, in most case $p<1$, under this assumption, we rewrite it to the form: $max(x, px)$, $(p<1)$. Since $p$ is a channel-wise parameter, it can be interpreted as a 1x1 depth-wise convolution regardless of the bias terms.

\subsubsection{Funnel condition}
FReLU adopts the same $max(\cdot)$ as the simple non-linear function.
For the condition part, FReLU extends it to be a 2D condition dependent on the spatial context for each pixel (see Fig. \ref{fig:framework}).
This is in contrast to most recent methods whose condition depends on the pixel itself (e.g. \cite{maas2013rectifier,he2015delving}) or the channel context (e.g. \cite{goodfellow2013maxout}).
Our approach follows the spirit of ReLU that uses a $max(\cdot)$ to obtain the maximum between $x$ and a condition.

Formally, we define the funnel condition as $\bT(x)$.
To implement the spatial condition, we use a \textbf{Parametric Pooling Window} to create the spatial dependency, specifically, we define the activation function:

\begin{equation}
f(x_{c, i, j})= max(x_{c, i, j}, \bT(x_{c, i, j}))
\end{equation}

\begin{equation}
\bT(x_{c, i, j}) = x^{\omega}_{c, i, j} \cdot p^{\omega}_{c}
\end{equation}

Here, $x_{c, i, j}$ is the input pixel of the non-linear activation $f(\cdot)$ on the $c$-th channel, at the 2-D spatial position $(i, j)$; function $\bT(\cdot)$ denotes the funnel condition, $x^{\omega}_{c, i, j}$ denotes a $k_h \times k_w$ \textbf{Parametric Pooling Window} centered on $x_{c, i, j}$, $p^{\omega}_{c}$ denotes the coefficient on this window which is shared in the same channel, and $(\cdot)$ denotes dot multiply.

\subsubsection{Pixel-wise modeling capacity}
Our definition of funnel condition allows the network to generate spatial conditions in the non-linear activations for every pixel.
The network conducts non-linear transformations and creates spatial dependencies $simultaneously$.
This is different from common practice which creates spatial dependency in the convolution layer and conducts non-linear transformations separately.
In that case, the activations do not depend on spatial conditions explicitly; in our case, with the funnel condition, they do.

\begin{figure}[t]
\centering
\includegraphics[height=2.5cm]{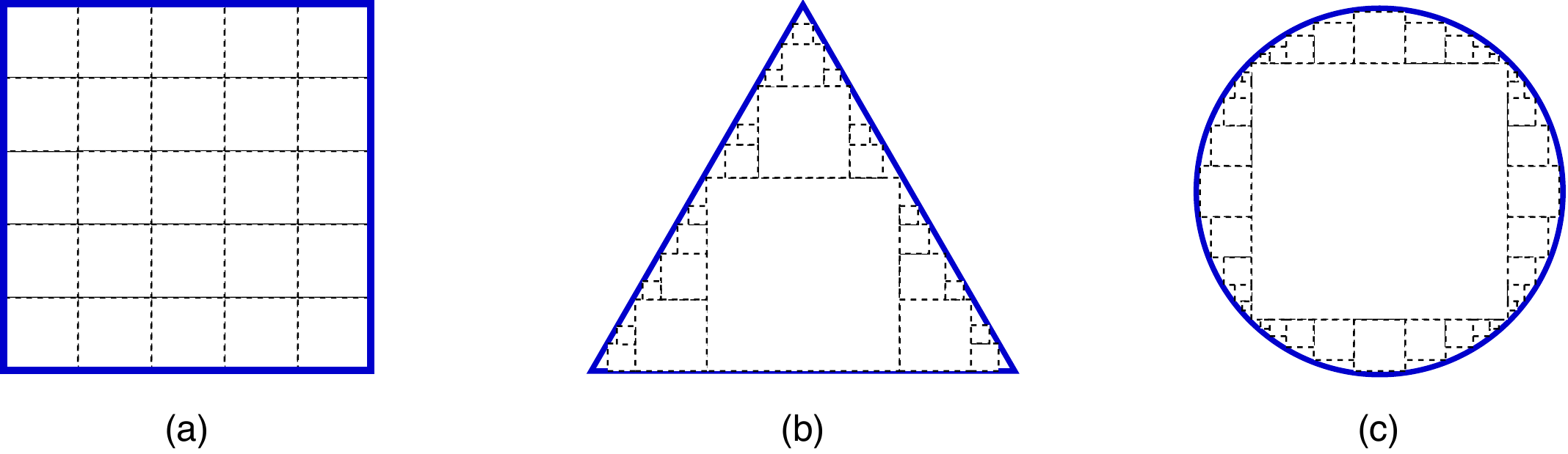}
\caption{Graphic depiction of how the per-pixel funnel condition can achieve \textit{pixel-wise modeling capacity}.
The distinct sizes of squares represent the distinct \textit{activate fields} of each pixel in the top activation layers.
(a) The normal activate field that has equal sizes of squares per-pixel, and can only describe the horizontal and vertical layouts. 
In contrast, the $max(\cdot)$ allows each pixel to choose \textit{looking around or not} in each layer, after enough number of layers, they have many different sizes of squares.
Therefore, the different sizes of squares can approximate (b) the shape of the oblique line, and
(c) the shape of an arc, which are more common natural object layouts.}
\label{fig:rf}
\end{figure}

As a result, the pixel-wise condition makes the network has a pixel-wise modeling capacity, the function $max(\cdot)$ gives per-pixel a choice between \textit{looking at the spatial context or not}.
Formally, consider a network $\{F_1, F_2, ..., F_n\}$ with $n$ FReLU layers, each FReLU layer $F_i$ has a $k\times k$ parametric window. For brevity, we only analyze the FReLU layers regardless of the convolution layers. Because the max selection between $1\times1$ and $k\times k$, each pixel after $F_1$ has a $activate\ filed$ set $\{1,1+r\}$ $(r=k-1)$. After the $F_n$ layer, the set becomes $\{1,1+r,1+2r,...,1+nr\}$, which gives more choices to each pixel and can approximate any layouts if $n$ is sufficiently large.
With many distinct sizes of the activate field, the distinct sizes of squares can approximate the shape of the oblique line and arc (see Fig. \ref{fig:rf}).
As we know, the layout of the objects in the images are usually not horizontal or vertical, they are usually in the shape of the oblique line or arc, therefore extracting the spatial structure
of objects can be addressed naturally by the pixel-wise modeling capacity provided by the spatial condition.
We show by experiments that it captures \textbf{irregular and detailed object layouts} better in complex tasks (see Fig. \ref{fig:seg}).

\subsection{Implementation Details}
Our proposed change is simple: we avoid the hand-designed condition in activations, we use a simple and effective spatial 2D condition to replace it.
The visual activation leads to significant improvements as shown in Fig. \ref{fig:result}.
We first change the ReLU activations in the classification task on the ImageNet dataset.
We use ResNet \cite{he2016deep} as the classification network and use the pre-trained network as backbones for other tasks: object detection and semantic segmentation.

All the regions $x^{\omega}_{c, i, j}$ in the same channel share the same coefficient $p^{\omega}_{c}$, therefore, it only adds a slight additional number of parameters.
The region represented by $x^{\omega}_{c, i, j}$ is a sliding window, the size is default set to a 3$\times$3 square, and we set the 2-D padding to be 1, in this case,

\begin{equation}
x^{\omega}_{c, i, j} \cdot p^{\omega}_{c} = \sum_{i-1 \leq h \leq i+1,\\ j-1 \leq w \leq j+1} x_{c, h, w} \cdot p_{c, h, w}
\end{equation}

\subsubsection{Parameter initialization}
We use the gaussian initialization to initialize the hyper-parameters.
Therefore we get the condition values close to zero, which does not change the origin network's property too much.
We also investigate the cases without parameters, (e.g. max pooling, average pooling), which do not show improvement.
That shows the importance of the additional parameters.

\subsubsection{Parameter computation}
We assume there is a $K'_h \times K'_w$ convolution with the input feature size of $C\times H \times W$ input, and the output size of $C \times H' \times W'$, then we compute the number of parameters to be $CCK'_hK'_w$, and the FLOPs (floating point operations) to be $CCK'_hK'_wHW$.
To this we add our funnel condition with window $K_h \times K_w$, the additional number of parameters is $CK_hK_w$, and the additional number of FLOPs is $CK_hK_wHW$.
We assume $K=K_h=K_w$,$K'=K'_h=K'_w$ for simplification. 

Therefore the original complexity of parameters is $O(C^2K'^2)$, after adopting FReLU, it becomes $O(C^2K'^2+CK^2))$;
and the original complexity of FLOPs is $O(C^2K'^2HW)$, after adopting the visual activation, it becomes $O(C^2K'^2HW + CK^2HW)$.
 Usually, $C$ is much larger than $K$ and $K'$, therefore the additional complexity can be negligible.
Actually in practice the additional part is negligible (more details in Table \ref{table:resnet}).
Moreover, the funnel condition is a $k_h \times k_w$ sliding window, and we implement it using the highly optimized depth-wise separable convolution operator followed with a BN \cite{ioffe2015batch} layer.

%

\section{Experiments}

\subsection{Image Classification}
\label{sec:cls}
To evaluate the effectiveness of our visual activation, first, we conduct our experiments on ImageNet 2012 classification dataset\cite{deng2009imagenet,russakovsky2015imagenet}, which comprises 1.28 million training images and 50K validation images.

Our visual activation is easy to adopt on the network structures, by simply changing the ReLU in the original CNN structure.
First, we evaluate the activation on different sizes of ResNet \cite{he2016deep}.
For the network structure, we use the original implementation.
Spatial dependency is important especially in the shallow layers, for the small 224$\times$224 input size, we replace the ReLUs in all the stages except the last stage, which has a small 7$\times$7 feature map size.
For the training settings, we use a batch size of 256, 600k iterations, a learning rate of 0.1 with linear decay schedule, a weight decay of 1e-4, and a dropout \cite{hinton2012improving} rate of 0.1.
We present the Top-1 error rate on the validation set.
For a fair comparison, we run all the results on the same code base.

\begin{table}[t]
\begin{center}
\caption{Comparisons with other effective activations \cite{he2015delving,ramachandran2017searching} on ResNets \cite{he2016deep} in ImageNet 2012. Image size 224x224. Single crop. We evaluate the Top-1 error rate on the test set.}
\label{table:resnet}
\begin{tabular}{llccc}
\hline
      Model      & Activation  & \#Params & FLOPs & Top-1 Err. \\ \hline
\multirow{4}{*}{ResNet-50} & ReLU & 25.5M  & 3.86G & 24.0    \\ 
              & PReLU & 25.5M  & 3.86G & 23.7    \\ 
              & Swish & 25.5M  & 3.86G & 23.5    \\ 
              & FReLU & 25.5M  & 3.87G & \textbf{22.4}    \\ \hline
\multirow{4}{*}{ResNet-101} & ReLU & 44.4M  & 7.6G & 22.8    \\ 
              & PReLU & 44.4M  & 7.6G & 22.7    \\ 
              & Swish & 44.4M  & 7.6G & 22.7    \\ 
              & FReLU & 44.5M  & 7.6G & \textbf{22.1}    \\ \hline
\end{tabular}
\end{center}
\end{table}

\subsubsection{Comparisons with scalar activations}
We conduct a comprehensive comparison on ResNets \cite{he2016deep} with different depths (e.g. ResNet-50, ResNet-101).
We take ReLU as the baseline and take one of its variants PReLU for comparison. Further, we compare our visual activation with the activation Swish \cite{ramachandran2017searching} searched by the NAS \cite{zoph2016neural,zoph2018learning} technique.
Swish has shown its positive influence on various model structures, comparing with many scalar activations.

Table \ref{table:resnet} shows the comparison, our visual activation still outperforms all of them with a negligible additional complexity.
Our visual activation improves 1.6\% and 0.7\% top-1 accuracy rates on ResNet-50 and ResNet-101.
It's remarkable that with the increase of model size and model depth, other scalar activations show limited improvement, while visual activation still has significant improvement.
For example, Swish and PReLU improve the accuracy of 0.1\% on ResNet-101, while visual activation increases still significantly on ResNet-101 with an improvement of 0.7\%.

\subsubsection{Comparison on light-weight CNNs}
Besides deep CNNs, we compare the visual activation with other effective activations on recent light-weight CNNs such as MobileNets \cite{howard2017mobilenets} and ShuffleNets \cite{ma2018shufflenet}.
We use the same training settings in \cite{ma2018shufflenet}.
The model sizes are extremely small, we use a window size of 1$\times$3+3$\times$1 to reduce the additional parameters.
Moreover, for MobileNet we slightly refine the width multiplier from 0.75 to 0.73 to maintain the model complexity. 
Table \ref{table:light} shows the comparison results on ImageNet dataset.
Our visual activation also boosts accuracy on light-weight CNNs.
ShuffleNetV2 0.5$\times$ can improve 2.5\% top-1 accuracy by only adding a slight additional FLOPs.

\begin{table}[t]
\begin{center}
\caption{Comparisons among other effective activations \cite{he2015delving,ramachandran2017searching} on light-weight CNNs (MobileNet \cite{howard2017mobilenets}, ShuffleNetV2 \cite{ma2018shufflenet}) in ImageNet 2012. Image size 224x224. Single crop. We evaluate the Top-1 error rate on the test set.}
\label{table:light}
\begin{tabular}{llccc}
\hline
      Model      & Activation  & \#Params & FLOPs & Top-1 Err. \\ \hline
\multirow{4}{*}{MobileNet 0.75 } & ReLU &  2.5M & 325M &  29.8  \\ 
              & PReLU &   2.5M & 325M &  29.6    \\ 
              & Swish &   2.5M & 325M &  28.9    \\ 
              & FReLU &   2.5M & 328M &  \textbf{28.5}    \\ \hline
\multirow{4}{*}{ShuffleNetV2} & ReLU &  1.4M & 41M &  39.6  \\ 
              & PReLU &  1.4M & 41M &  39.1   \\ 
              & Swish &  1.4M & 41M &  38.7   \\ 
              & FReLU &  1.4M & 45M & \textbf{37.1}    \\ \hline
\end{tabular}
\end{center}
\end{table}

\subsection{Object Detection}
To evaluate the generalization performance of visual activation on different tasks, we conduct object detection experiments on COCO dataset \cite{lin2014microsoft}.
The COCO dataset has 80 object categories. We use the $trainval35k$ set for training and use the $minival$ set for testing.

\begin{table}
\begin{center}
\caption{Comparisons of different activations in COCO \textbf{object detection}. We use ResNet-50 \cite{he2016deep} and ShuffleNetV2 (1.5$\times$) \cite{ma2018shufflenet} with different activations as the pre-trained backbones. We use the RetinaNet \cite{lin2017focal} detector.}
\label{table:det}
\begin{tabular}{lccccccccc}
\hline
      Model      & Activation  & \#Params & FLOPs & mAP & $AP_{50}$ & $AP_{75}$ & $AP_{s}$ & $AP_{m}$ & $AP_{l}$ \\ \hline
\multirow{3}{*}{ResNet-50} & ReLU & 25.5M  & 3.86G & 35.2 &53.7&37.5&18.8&39.7&48.8   \\ 
              & Swish & 25.5M  & 3.86G & 35.8 &54.1&38.7&18.6&40.0&49.4   \\ 
              & FReLU & 25.5M  & 3.87G & \textbf{36.6} &\textbf{55.2}&\textbf{39.0}&\textbf{19.2}&\textbf{40.8}&\textbf{51.9}    \\ \hline
\multirow{3}{*}{ShuffleNetV2} & ReLU & 3.5M  & 299M & 31.7  &49.4&33.7&15.3&35.1&45.2  \\ 
              & Swish & 3.5M  & 299M & 32.0  &49.9&34.0&16.2&35.2&45.2  \\ 
              & FReLU & 3.7M  & 318M & \textbf{32.8}  &\textbf{50.9}&\textbf{34.8}&\textbf{17.0}&\textbf{36.2}&\textbf{46.8}  \\ \hline

\end{tabular}
\end{center}
\end{table}

We present the result on RetinaNet \cite{lin2017focal} detector.
For a fair comparison, we train all the models in the same code base with the same settings.
 We use a batch size of 2, a weight decay of 1e-4 and a momentum of 0.9. We use anchors for 3 scales and 3 aspect ratios and use a 600-pixel train and test image scale.
For the backbone, we use the pre-trained model in Section \ref{sec:cls} as a feature extractor, and compare the generality among different activations.

Table \ref{table:det} shows the comparison among different activations.
The comparison shows that our visual activation increases 1.4\% mAP comparing to the ReLU backbone, and increases 0.8\% mAP comparing to the Swish backbone.
It is worth mentioning that, on all the small, medium, and large objects, FReLU outperforms all the other counterparts significantly.

We also show the comparison on the light-weight CNNs.
As the comparison of ResNet-50, we use pre-trained ShuffleNetV2 backbones adopted with different activations.
We mainly compare FReLU with ReLU and the effective activation Swish \cite{ramachandran2017searching}.
Table \ref{table:det} shows visual activation also outperforms much better than ReLU and Swish backbones, to which it improves 1.1\% mAP and 0.8\% mAP respectively.
Moreover, it increases the performance of all the sizes of objects.

\begin{table}[t]
\begin{center}
\caption{Comparisons on the \textbf{semantic segmentation} task in CityScape dataset. We use the PSPNet \cite{zhao2017pyramid} as the the framework and use the ResNet-50 \cite{he2016deep} as backbone. The pre-trained backbones are from Table \ref{table:resnet}.}
\label{table:seg}
\begin{tabular}{lccc}
\hline
  & ReLU & Swish\cite{ramachandran2017searching} & FReLU \\ \hline
mean\_IU &  77.2 & 77.5   & \textbf{78.9}  \\ \hline
road	 & 98.0 	 & 98.1 	 & 98.1 \\ 
sidewalk	 & 84.2 	 & 85.0 	 & 84.7 \\ 
building	 & 92.3 	 & 92.5 	 & 92.7 \\ 
wall	 & 55.0 	 & 56.3 	 & 59.5 \\ 
fence	 & 59.0 	 & 59.6 	 & 60.9 \\ 
pole	 & 63.3 	 & 63.6 	 & 64.3 \\ 
traffic light	 & 71.4 	 & 72.1 	 & 72.2 \\ 
traffic sign	 & 79.0 	 & 80.0 	 & 79.9 \\ 
vegetation	 & 92.4 	 & 92.7 	 & 92.8 \\ 
 \hline
 \end{tabular}\hskip 2em
 \begin{tabular}{lccc}
 \hline
  & ReLU & Swish & FReLU \\ \hline
terrain	 & 65.0 	 & 64.0 	 & 64.5 \\ 
 sky	 & 94.7 	 & 94.9 	 & 94.8 \\ 
 person	 & 82.1 	 & 83.1 	 & 83.2 \\ 
 rider	 & 62.3 	 & 65.5 	 & 64.7 \\ 
 car	 & 95.1 	 & 94.8 	 & 95.3 \\ 
 truck	 & 77.7 	 & 70.1 	 & 79.8 \\ 
 bus	 & 84.9 	 & 84.0 	 & 87.8 \\ 
 train	 & 63.3 	 & 68.8 	 & 74.6 \\ 
 motorcycle	 & 68.3 	 & 69.4 	 & 69.8 \\ 
 bicycle	 & 78.2 	 & 78.4 	 & 78.7 \\
 \hline

\end{tabular}
\end{center}
\end{table}

\subsection{Semantic Segmentation}
We further present the semantic segmentation results on CityScape dataset \cite{cordts2016cityscapes}.
The dataset is a semantic urban scene understanding dataset, contains 19 categories.
It has 5,000 finely annotated images, 2,975 for training, 500 for validation and 1525 for testing.

\begin{figure}[t]
\centering
\includegraphics[height=7cm]{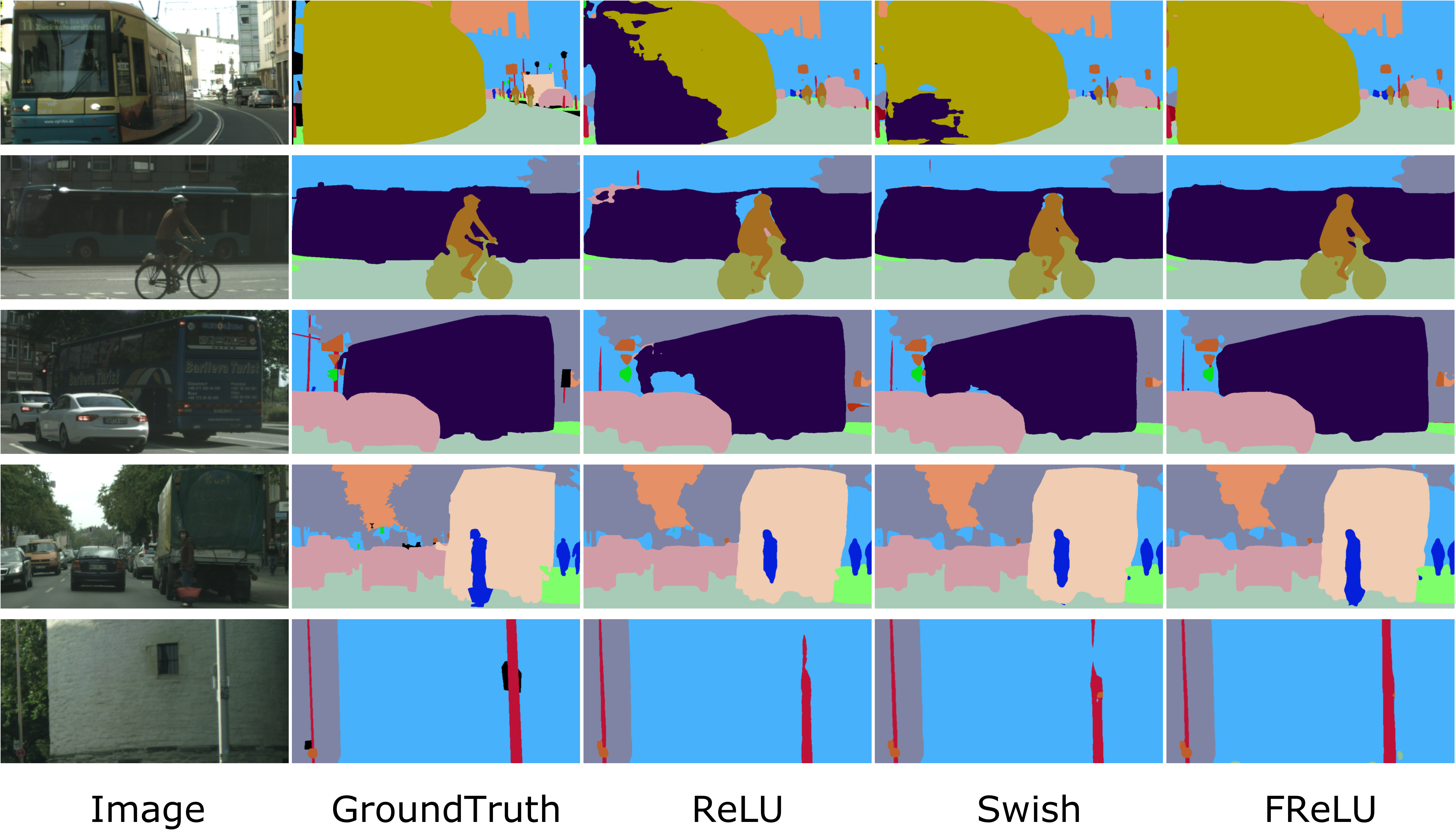}
\caption{\textbf{Visualization of semantic segmentation} on ResNet-50\cite{he2016deep}-PSPNet\cite{zhao2017pyramid} with different activations in backbone. We clip the CityScape images to make the differences more clear (better view enlarge images). FReLU has better long-range (large or slender objects) and short-range (small objects) understandings due to its better context capturing capacity.
It captures irregular and detailed object layouts in complex cases much better. 
We note that modern frameworks are finely optimized with ReLU, however, it has obvious improvements by only changing the backbones, thus having the potential for further gains if redesign the frameworks for the visual activation.
}
\label{fig:seg}
\end{figure}

We use the PSPNet \cite{zhao2017pyramid} as the segmentation framework, for the training settings we use the poly learning rate policy \cite{chen2017deeplab} where the base is 0.01 and the power is 0.9, we use a weight decay of 1e-4, and 8 GPUs with a batch size of 2 on each GPU.

To evaluate the generality of the previous pre-trained models in Section \ref{sec:cls},
we use the pre-trained ResNet-50 \cite{he2016deep} backbone models with different activations, we compare FReLU with Swish and ReLU respectively.

In Table \ref{table:seg}, we show the comparison with scalar activations.
From the result, we observe that our visual activation outperforms the ReLU and the searched Swish 1.7\% and 1.4\% mean\_IU, respectively.
Moreover, our visual activation has significant improvements in both large and small objects, especially on categories such as 'train', 'bus', 'wall', etc.

For better visualization of the improved performance, Fig. \ref{fig:seg} shows the predict results on the testing dataset.
It shows that by only changing the backbone activations, the results have obvious improvement.
The boundaries of both the large and the small objects are well-segmented because the pixel-wise modeling capacity can handle both global and detailed regions (see Fig. \ref{fig:rf}).
We note that the modern recognition frameworks are finely designed with the ReLU activation, therefore the visual activation still has great potential for further improving the results, which is beyond the focus of this work.

\section{Discussion}
The previous sections demonstrate the optimum performance comparing with other effective activations.
To further investigate our visual activation, we conduct ablation studies.
We first discuss the properties of the visual activation, then we discuss the compatibility with existing methods.

\subsection{Properties}

Our funnel activation mainly has two components: 1) funnel condition, and 2) $max(\cdot)$ non-linearity.
Separately, we investigate the effect of each component.


\begin{table}[t]
\begin{minipage}{0.55\linewidth}
\centering

\caption{Ablation on the different \textbf{spatial condition manners}, and the different \textbf{non-linear manners}. The experiments are conducted on ResNet-50 \cite{he2016deep}. 
Model A, B, C compare different visual conditions with/without parameters.
Model D replaces $max$ with $sum$, to this we add a ReLU, or it will not converge.
Model E separates and evaluates the performance of the spatial condition itself.
DW(x) represent the 3x3 depth-wise separable convolution.}
\label{table:dwconv}
\begin{tabular}{clc}
\hline
      Model      & Activation  & Top-1 Err. \\ \hline
A  & Max(x, ParamPool(x))  &  \textbf{22.4}  \\ 
 B  & Max(x, MaxPool(x)) &  24.4   \\ 
 C  & Max(x, AvgPool(x)) &   24.5  \\ \hline
 A  & Max(x, ParamPool(x))  &  \textbf{22.4}  \\ 
 D  & Sum(x, ParamPool(x))  &   23.6  \\ 
 E  & Max(DW(x), 0)  & 23.7    \\ \hline

\end{tabular}

\end{minipage}
\hfill
\begin{minipage}{0.4\linewidth}  
\centering
\caption{Ablation on different normalization methods after the spatial condition layer. We adopt Batch Normalization (BN) \cite{ioffe2015batch}, Layer Normalization (LN) \cite{ba2016layer}, Instance Normalization (IN) \cite{ulyanov2016instance} and Group Normalization (GN) \cite{wu2018group} after the spatial condition layer which is implemented by depth-wise convolution. ImageNet results on ShuffleNetV2 0.5$\times$.}
\label{table:norm}

\begin{tabular}{cc}
\hline
 Normalization & Top-1 Err.\\ \hline
  - &37.6 \\
  BN &37.1 \\
  LN &36.5 \\
  IN&38.0 \\
  GN&36.5 \\ \hline

\end{tabular}

\end{minipage}
\end{table}

\subsubsection{Ablation on the spatial condition}
First, we compare the different manners of the spatial condition.
Besides the manner of parametric pooling that we used, to investigate the importance of the additional parameters, we compare other pooling manners without additional parameters, they are max pooling and average pooling.
We simply replace the parametric pooling with the other two non-parametric manners and evaluate the results on the ImageNet dataset.

Table \ref{table:dwconv} (A, B, C) shows the importance of the parametric pooling.
Without additional parameter, the results decrease more than 2\% top-1 accuracy, even perform worse than the baseline that does not use spatial condition.
Table \ref{table:norm} shows the comparison of different normalization after the spatial condition.

\subsubsection{Ablation on the non-linearity}
Second, we also compare the use of non-linearity.
In our method, we use the $max(\cdot)$ function to perform the non-linearity, $simultaneously$ capturing visual dependency.
In contrast, we compare with the manners that $separately$ capture visual dependency and non-linearity.

For the spatial context capturing, we use two manners: 1) use the parametric pooling as before, then linearly add up with the original feature, 2) simply add a depth-wise separable convolution layer.
For the non-linear transformation, we use the ReLU function.
Table \ref{table:dwconv} (A,D,E) show the results.
Comparing with the baseline, the spatial context itself improves about 0.3\% accuracy, but together as the non-linear condition in our method, it further increases more than 1\%.
Therefore, performing the spatial dependency and non-linearity $separately$ has not an ideal effect as doing them $simultaneously$.

\subsubsection{Ablation on the window size}
In the parametric pooling window, the size of the window decides the size of the area each pixel $looks$.
We simply change the window size in the funnel condition and compare different sizes among \{$1\times 1, 3\times 3, 5\times 5, 7\times 7$\}.
The case of $1 \times 1$ does not have the spatial condition and it is the case of PReLU since the parameter value is smaller than 1.
Table \ref{table:window} shows the comparison results.
We conclude that $3 \times 3$ is the best choice.
The larger window sizes also show benefits but do not outperform 3$\times$3.

Further, we consider the case using an irregular window instead of squares.
We use multiple windows of sizes $1\times3$ and $3\times1$, we consider to use the sum and max of them as the condition.
Table \ref{table:window} \{B,E,F\} show the comparison.
The results show that irregular window sizes also have the optimum performance since they have a more flexible pixel-wise modeling capacity (Fig. \ref{fig:rf}).


\subsection{Compatibility with Existing Methods}
To adopt the new activation into the convolutional networks, we have to choose which layers, and which stages to adopt.
Moreover, we also investigate the compatibility with existing effective approaches such as SENet.


\begin{table}[t]
\begin{minipage}{0.49\linewidth}
\centering

\caption{Ablation on the \textbf{window size}. We simply change the window size in the funnel condition.
We evaluate the top-1 error rate on ImageNet dataset using the ResNet-50 \cite{he2016deep} structure.}
\label{table:window}
\begin{tabular}{ccc}
\hline
       Model&      Window size      &  Top-1 Err. \\ \hline
A&1$\times$1  &   23.7 \\ 
B&3$\times$3  &  \textbf{22.4}  \\ 
C&5$\times$5  &   22.9  \\ 
D&7$\times$7  &    23.0 \\ 
E& Sum(1$\times$3,3$\times$1)  &   22.6 \\ 
F&Max(1$\times$3, 3$\times$1)  &  \textbf{22.4}   \\ \hline

\end{tabular}

\end{minipage}
\hfill
\begin{minipage}{0.49\linewidth}  
\centering
\caption{Ablation on \textbf{different layers}. We replace the ReLU with FReLU after the 1$\times$1 convolution and the 3$\times$3 convolution.
Results are performed on ResNet-50 \cite{he2016deep} and MobileNet \cite{howard2017mobilenets}.}
\label{table:layer}

\begin{tabular}{lccc}
\hline
   &   1$\times$1 conv.     & 3$\times$3 conv. & Top-1 Err. \\ \hline
 \multirow{3}{*}{ResNet-50} &\checkmark &  &    22.9    \\ 
& & \checkmark &   23.0     \\ 
& \checkmark   & \checkmark &   \textbf{22.4}    \\ \hline
 \multirow{3}{*}{MobileNet} &\checkmark &  &     29.2  \\ 
& & \checkmark &    29.0   \\ 
& \checkmark   & \checkmark &    \textbf{28.5}    \\ \hline

\end{tabular}

\end{minipage}
\end{table}


\begin{table}[t]
\begin{minipage}{0.53\linewidth}
\centering

\caption{Ablation of visual activation on \textbf{different stages} (Stage \{2-4\} in ResNet-50 \cite{he2016deep}). In each stage we replace each ReLU with our visual activation.
The results are the top-1 error rates on ImageNet. Image size 224x224. }
\label{table:stage}

\begin{tabular}{cccc}
\hline
      Stage 2      & Stage 3  & Stage 4  & Top-1 Err. \\ \hline
\checkmark &  &   &  23.1   \\ 
 & \checkmark &   &   23.0  \\ 
  &  & \checkmark  &   23.3  \\ 
\checkmark & \checkmark &   &  22.8  \\ 
 & \checkmark &  \checkmark &   23.0  \\ 
\checkmark & \checkmark &  \checkmark &  \textbf{22.4}    \\ \hline

\end{tabular}

\end{minipage}
\hfill
\begin{minipage}{0.43\linewidth}  
\centering
\caption{Ablation comparisons of the compatibility between FReLU and SENet \cite{hu2018squeeze} on ResNet-50 \cite{he2016deep}.
The results are the top-1 error rates on ImageNet. Image size 224x224. Single crop.}
\label{table:se}

\begin{tabular}{lccc}
\hline
      Model       & \#Params & FLOPs & Top-1 \\ \hline
 ReLU & 25.5M  & 3.9G & 24.0    \\ 
 FReLU & 25.5M  & 3.9G & 22.4 \\
 ReLU+SE & 26.7M & 3.9G & 22.8 \\
 FReLU+SE & 26.7M  & 3.9G & \textbf{22.1}      \\ \hline

\end{tabular}


\end{minipage}
\end{table}

\subsubsection{Compatibility with different convolution layers}
First, we compare the positions after different convolution layers. 
That is, we investigate the effect of FReLU in different positions after 1$\times$1 and 3$\times$3 convolutions.
We conduct experiments on ResNet-50 \cite{he2016deep} and ShuffleNetV2 \cite{ma2018shufflenet}.
We replace the ReLU after the 1$\times$1 convolution and the 3$\times$3 convolution and observe the improvement.
Table \ref{table:layer} shows the results, in the bottleneck of the above two networks.
From the results, we can see that the improvements on different layers are comparable, and it has the optimum performance when adopting both of them.

\subsubsection{Compatibility with different stages}
Secondly, we investigate the compatibility with different stages in the CNN structures.
The visual activations are important especially on the layer with high spatial dimensions.
For the classification network whose shallow layers have larger spatial dimensions and deeper layers have large channel dimensions, there may be differences when we apply visual activations on different stages.
For Stage 5 of ResNet-50 with 224x224 input, it has a relatively small 7x7 feature size, which mainly contains channel dependency instead of spatial dependency.
Therefore, we adopt visual activations on Stage \{2-4\} on ResNet-50, as Table \ref{table:stage} shows.
The results reveal that adopting the shallow layers has a larger effect, while a deeper layer has a smaller effect. 
Moreover, adopting FReLU on all of them has the optimum top-1 accuracy.

%
%
%
%
%

\subsubsection{Compatibility with SENet}
At last, we compare the performance with SENet \cite{hu2018squeeze} and show the compatibility with it.
Without the complex advances in CNN architecture, it achieves significant improvements on all the three vision tasks, simply together with the regular convolution layers.
We further compare visual activation with recent effective attention module SENet, since SENet is one of the most effective attention modules recently.

Table \ref{table:se} shows the result,
although SENet uses an additional block to enhance the model capacity, 
it is remarkable that the simple visual activation even outperforms SENet.
We also wish the visual activation we proposed can co-exist with other techniques, such as the SE module.
We adopt the SE module on the last stage in ResNet-50 to avoid overfitting.
Table \ref{table:se} also shows the co-existence between FReLU and SE module.
Together with SENet, funnel activation improves 0.3\% accuracy further.

\section{Conclusions}

In this work, we present a funnel activation specifically designed visual tasks, which easily captures complex layouts using the pixel-wise modeling capacity.
Our approach is simple, effective, and finely compatible with other techniques, that provides a new alternative activation for image recognition tasks.
We note that ReLU has been so influential that many state-of-the-art architectures have been designed for it, however, their settings may not be optimal for the funnel activation.
Therefore, it still has a large potential for further improvements.

\subsubsection{Acknowledgements}
This work is supported by The National Key Research and Development Program of China (No. 2017YFA0700800) and Beijing Academy of Artificial Intelligence (BAAI).

\clearpage
%
%
\bibliographystyle{splncs04}
\bibliography{egbib}
\end{document}